# Can machines be uncertain?

Luis Rosa (Washington University in St. Louis)

## 1. Introduction

Can AI systems be *uncertain* about things? If yes, then *how* do they realize states of uncertainty?

These issues matter, for we do want AI systems to be able to be uncertain, as opposed to having an answer to everything. We do not want them to 'jump to conclusions', for example. On the other hand, we do not want them to be uncertain about issues they have enough information to settle. We want them to be uncertain just in case *they should* be uncertain.

But saying that only makes sense if AI systems are indeed capable of being uncertain.

So, are they capable of that, again? If so, then how?

Philosophers have already debated the issue whether AI systems can have *propositional* attitudes, such belief and intention. As recently remarked by Chalmers (2025), however, the issue only becomes tractable (and less trivial) when we don't already require consciousness or even the possession of a mind for having such attitudes.

We may countenance an AI system as having propositional attitudes because it manipulates representations of reality in certain ways, say.[1] What would be left for us to determine, then, is what kind of role more exactly those representations have to play within the system, and what kinds of further processes the AI system needs to perform on the basis of those representations, in order for that system to count as having this or that propositional attitude. For example, what kinds of representations does the system need to have, and what operations does it need to perform with those representations, in order to count as *believing* something, as opposed to, say, just having a *high credence* in it, or merely *supposing* that it is true?[2]

We may also countenance an AI system as having propositional attitudes because it *behaves as if* it had them, or because we can make good sense of its behavior by ascribing such attitudes (this may *either* be seen as an alternative to *or* as a complement to the previous proposal).[3] What would be left for us to determine, then, is what kinds of behavior go with which propositional attitudes, or what kinds of behavior are best made sense of through which ascriptions of attitudes. And here we may work with a broader or a narrower notion of behavior, depending on the kind of system we are looking at.

---

[1] That would assume some form of functionalism about propositional attitudes (see Lewis 1972, Putnam 1975). Though note that we may decide to work with a *technical* or conceptually engineered notion of these attitudes, without presupposing that the *ordinary* notions of propositional attitudes from folk psychology are to be fully analyzed in this way, too.

[2] For an overview of the literature on the relationship between belief and credence, see Jackson (2020).

[3] See Dennett (1971) for a congenial approach, even as it applies to biological cognitive systems.

It is in this same spirit that I raise the questions about uncertainty in AI systems from above.

The goal is to establish whether and when we can countenance different AI systems as *being uncertain* about different things—but in raising that question I do not already presuppose that those systems must have full-blown minds or consciousness in order to be uncertain.

If having a full-blown mind or consciousness were needed for uncertainty, then we would *first* have to settle the question whether AI systems can have full-blown minds or consciousness, and only then address the question whether AI systems can be uncertain. But at this point we want to sidestep questions about the possibility of full-blown mentality in machines and machine consciousness, and focus more directly on the functional and behavioral aspects of uncertainty.

Importantly, what I am after here are ascriptions of uncertainty *to the AI systems themselves*, and not merely to the information or knowledge available to them—as when we say that *their data* itself is uncertain, because it is inconclusive regarding some specific issue.

Uncertainty *in the information* stored in the system is not sufficient. I also want to characterize *the cognitive system itself* as being uncertain about things, as having an attitude of uncertainty. In other words, I am interested in ascriptions of *subjective* uncertainty, or uncertainty at the level of the system's opinions or stances, and not merely ascriptions of *epistemic* uncertainty (uncertainty in the information or knowledge possessed by the system).[4]

To illustrate the difference, consider non-artificial cognitive systems such as ourselves. It may happen that my total evidence leaves it open whether there are intelligent aliens in some distant planet—but I am 100% sure that there are such creatures (this is luckily a hypothetical example). In this case, there is epistemic uncertainty but not subjective uncertainty in my cognitive system, seeing as my total evidence doesn't resolve the question, though I cognize as if it did. Mismatches the other way around are possible, too. I can be uncertain whether I am capable of learning Italian, despite having conclusive evidence that I am perfectly capable of learning it. In this case, there is subjective but no epistemic uncertainty in my cognitive system.

Similarly, we want to know whether AI systems themselves can be uncertain about things, and not only whether the information they store harbors uncertainty. This distinction between epistemic and subjective uncertainty will prove useful in the discussion below.

## 2. Justifying the question further

The reader may wonder why we should raise the question about *uncertainty* in AI systems in particular, and not simply join the already established literature on whether AI systems can have propositional attitudes in general.

In the previous section, I have already given some indication as to why it is important to be concerned with uncertainty in particular—we *want* AI systems to be capable of uncertainty, and not only belief or intention. But the full justification for this focus is also backed up by the observation that some states of uncertainty are essentially *interrogative* attitudes, irreducible to *propositional* attitudes. Accordingly, their contents are questions and not propositions.

---

[4] This distinction is analogous to the one between subjective and epistemic *certainty* from Stanley (2008).

Here is a case to illustrate the point, however briefly.

Maria is a young student who is learning facts about the solar system with her teacher. At this point, Maria knows that there are eight planets in the solar system—but she doesn't know what those planets are yet. She doesn't know, for example, that Mercury is one of those planets, and so is Venus, and Earth, etc. Her teacher then poses the following question to her and the other students: What is the largest planet in the solar system? Maria understands the question perfectly well, but she doesn't have an answer to it. In fact, she doesn't even know what the *candidate* answers to that question are yet (that Mercury is the largest planet in the solar system, that Venus is the largest planet in the solar system, etc.) Maria doesn't have *any* propositional attitudes toward those propositions, because she cannot even think with them yet. She doesn't yet have a concept of *Mercury*, for example, and she doesn't know that 'Mercury' names a planet in the solar system (similarly for the other planets). Again, that doesn't mean that Maria doesn't understand the question. She understands every word in the interrogative 'What is the largest planet in the solar system?', and she is competent with its grammar, too. Maria knows what an answer to that question would look like—it would have the form *x is the largest planet in the solar system*. It is only that she doesn't know what those *x*s could be yet.

In this example, Maria is uncertain about *what the largest planet in the solar system is*. But her uncertainty doesn't boil down to her holding any propositional attitudes toward the answers to that question (e.g., credences whose propositional contents are those answers).

Could we say that Maria's uncertainty reduces rather to some metacognitive propositional state—e.g. a belief to the effect that *she doesn't know* what the largest planet in the solar system is? It is more or less clear that this strategy won't work in general, either. As already remarked by others, small children and non-human animals are also capable of having interrogative/questioning attitudes, even when they don't yet have a theory of mind or metacognition skills to harbor such sophisticated propositional attitudes (regarding what they themselves know or do not know, what they believe or do not believe, etc.).[5] A creature can be uncertain about something without believing that it doesn't know the answer to the question, or anything of the sort. In reaching a state of uncertainty, it may be simply processing information *about the surrounding environment*, without processing any information *about itself.*

We don't need to get bogged down in these details now. The general point is just that some states of uncertainty are fundamentally interrogative, in that they take *questions* for their content and are not reducible to *propositional* attitudes. Accordingly, the mere fact that an AI system has propositional attitudes doesn't guarantee that it is uncertain about anything—or at least not *always*.

---

[5] See for example Carruthers (2018), who is mostly concerned with *curiosity*. Curiosity involves uncertainty, of course, though one can be uncertain about something without being curious about it. But Carruthers' point applies to sheer uncertainty too (if one can $\phi$ *and* $\psi$ without $\sigma$-ing, then one can $\phi$ without $\sigma$-ing). On the notion of interrogative attitudes, see also Friedman (2017), Willard-Kyle (2023) and Rosa (2024).

And that again adds motivation to ask the more specific question about the possibility of uncertainty in AI systems, and not only propositional attitudes in general.

Another motivation stems from the fact that intelligence itself seems to require the capacity for uncertainty. In fact, a cognitive system that jumps to conclusions instead of being uncertain will in many cases opt for the less optimal course of action, thus making it harder for us to countenance it as intelligent. Uncertainty even has survival and adaptivity value. A creature who is capable of being uncertain whether there is a predator inside the cave is more likely to adapt and survive than one that can only have 'black or white' states of mind, for example, necessarily either believing that there is a predator inside the cave (thus not testing whether this is true in any way, but rather just running with it) or believing that there isn't one (thus risking being devoured by a predator). Intelligent systems have uncertainty when they should, and they strike a balance between opposite states of mind that are risky for them to adopt—because their evidence leaves the question open, or because there is too much at stake in making up their mind either way.

Since uncertainty is an important ingredient of intelligence, artificial *intelligence* must feature artificial *uncertainty*. Thus it is that the very search for artificial intelligence leads us to the questions about uncertainty in AI systems raised above.

## 3. AI Systems

The issue here is whether AI systems are capable of being uncertain and, if so, then *how* they realize states of uncertainty.

One would expect the answers to those questions to be tailored to *types* of AI systems—for different architectures will presumably realize states of uncertainty in different ways. Additionally, uncertainty can be ascribed to AI systems at different *levels* of interpretation. Before addressing the latter issue, however, let me first establish a minimal taxonomy of AI systems.

AI systems are typically divided in three general kinds: (a) Symbolic AI systems, (b) Connectionist systems, and (c) Hybrid systems containing elements of both, symbolic and connectionist AI. The contrast between them lies on the nature of *knowledge representation* and the *type of information processing* that takes place in each of them.[6]

Systems of type (a) operate with explicitly stated rules that tell them which strings of symbols to output in response to which inputs, given their model of the world/environment. The inputs, outputs and internal models are all symbolically encoded (Newell and Simon 1972). The target rules may have a logical, probabilistic or heuristic character. Either way, they deploy or refer to the same formal language that is used to represent the information that comes in and out of the system.

For example, an expert symbolic AI system may be used to issue medical diagnoses by operating on inputs of the forms *fever*($x$) and *coughs*($x$) to output tokens of the form *flu*($x$)—where *fever*($x$) reads *x has fever*, and similarly for the other formulas (this example is based on the expert system MYCIN by Shortliffe and Buchanan 1975). In this case, the inputs are sets of atomic declarative formulas, and the output is also an atomic declarative formula. There will then be a rule or set of rules in the system's algorithm that make the step between the former formula types and the latter formula type possible—and those rules deploy or refer to those very formula types.

[6] See Mitchell (2019) for a general introduction, and Bermúdez (2023, Ch. 8) for an overview of these AI architectures.

Systems of type (b), in contrast, operate with Artificial Neural Networks or ANNs (called this way because they are inspired by real/biological networks of neurons.)[7] The inputs and outputs of a neural network are not declarative sentences or formulas like *fever*($x$) and *flu*($x$)—but rather vectors of activation or ordered tuples of numbers, which can in turn stand for many different things (including sentences). The activations are activations of the ANNs *units* or artificial neurons, which are in turn organized in different layers—there is an input layer of units, hidden layers (if any), and an output layer of units.

Different vectors of activation in an ANNs's input layer of units convey different bits of information. For example, the input units of a *visual classifier* may convey information about the colors that fill each pixel of an image—so the activation vector of the input layer will have the form $<x_1,\ldots,x_n>$, where each $x_i$ is a code for the color that fills pixel $i$ of the target image. This information will then be processed further and cause activations in other layers of the network. These other activations are a function of the signals sent by units from the previous layers and the weights of the connections between them (those signals may be excitatory/positive, or inhibitory/negative, depending on the valence of those weights and the activation values of those previous units). The ANN will eventually output a final vector, which will in this case stand for some category—say, a vector that stands for the category of *bears*. The classifier will by right in this case only if the image encoded by the input layer is indeed the image of a bear. If it isn't, this error or discrepancy can be used to modify the weights of the network, until it gets the category of the target object right (see LeCun, Bengio and Hinton 2015).

Neural networks encode information in a *distributed* manner. They do that through the weights of the connections between their units, in combination with what their vectors of activation stand for. That information doesn't have to be explicitly stated by any formulas or sentences stored in the system (or even *follow from* formulas or sentences used by it), as in a purely symbolic AI system.

To bring out this contrast, compare the symbolic AI system for diagnosing diseases mentioned above to a connectionist counterpart. Let us say that the symbolic AI system will output a flu diagnosis when the patient satisfies a certain complex open formula $\phi$, translating a conjunction of ascriptions of symptoms (*x has fever and x coughs and x has headache*…). So the system will be operating with some explicitly stated generalization that deploys that open formula, such as: *if* $\phi(x)$ *then flu*($x$). But the connectionist AI system that performs that very same task (diagnosis) does not encode that information in that way. Rather than that, its patterns of activation are such that, whenever the information that $\phi(x)$ is encoded by its input layer (through its vector of activations), the information that *flu*($x$) is encoded by its output layer.[8] And that pattern of activation hinges again on how the weights of the connections between the units within the network are set.

[7] Though note that there are many discrepancies between the units of an ANN and a real neuron—see Bermúdez (2023, Ch. 5) for more on this.

In such a case, we can say that the *whole network* contains the information that if a patient has all those symptoms ascribed by $\phi$, then that patient has flu. But it doesn't do that through any *formula* that conveys that information, again, or even a set of formulas from which that formula follows. That information is only *implicit* in the structure of the network.

This is but an oversimplified contrast between symbolic and connectionist AI, of course. But it already illustrates how information is stored and processed in different ways by these two general types of systems.

Hybrid systems, as the name says, are a mix of symbolic and connectionist AI.[9] Large Language Models (LMMs) are a case in point, in that they store information about linguistic regularities through the weights and connections of their massive transformer networks—though at the same time they can also integrate symbolically encoded rules that allow them to perform reasoning from/to sentence or formulas (see Xiong et al. 2024, Sullivan and Elsayed 2024 for more).

With these distinctions in mind, we can now raise more specific versions of the questions raised at the beginning of this investigation. Can *symbolic AI* systems be uncertain? Can *connectionist AI* systems be uncertain? What about hybrid architectures? And how is uncertainty realized in each of these types of system (if at all)? I turn to these more specific questions now.

## 3. Uncertainty in symbolic AI systems

There are different ways in which states of uncertainty can be realized in symbolic AI system, at least in principle. They vary under two dimensions, namely, the *type* of uncertainty they realize (3.1, 3.2) and the *level* at which the uncertainty is realized (3.3).

### 3.1 Probabilistic uncertainty realized symbolically

One type of uncertainty that can be realized in symbolic AI systems is a probabilistic type of uncertainty. It stems from the fact that such systems can have different *degrees of confidence* toward certain claims.

For example, the rules implemented in a symbolic AI system may generate a 90% degree of confidence that a patient has a certain disease $D$, given information about their symptoms. And that degree of confidence can in turn be realized in different ways. It can be realized through a symbolic structure of the form $<D(a), 0.9>$, say, where $D(a)$ says that patient $a$ has disease $D$, and 0.9 marks the system's subjective probability for that claim. The system's degrees of confidence here correspond to it's own rule-based information processing outputs, in this case, ordered pairs of declarative sentences and real numbers between 0 and 1 (see Pearl 1988 for an overview of symbolic probabilistic reasoning). The system then *stores* such ordered pairs in its memory and integrates it into its internal model of the world. Quite generally, we can say that the system has at least *some* uncertainty regarding whether *p* when it outputs or stores a pair <*p*, *r*>, where *r* lies strictly between 0 and 1, for any *p*.

---

[8] For example, in the simplest possible way in which this can be implemented, the ANN's input layer will have $n$-many units, each responsible for registering the presence or absence of a given symptom: for each unit, its activation will be 1 if the symptom is present, 0 if absent.

[9] An early hybrid architecture is Anderson's (1976) ACT-R architecture, which attempts to emulate brain function when it comes to taking in information from the environment and acting to modify it in a way as to bring about one's goals.

That is just one possible way in which probabilistic uncertainty can be realized in symbolic AI systems. Another is that the system's degrees of confidence are merely *comparative*, corresponding rather to formulas like $Pr(p) > Pr(q)$ and $Pr(p) = Pr(q)$, say, which can both constitute the outputs of its algorithmic processes and be stored in its memory/integrated into its internal model. We can then say the system has at least some uncertainty regarding whether *p* when there is a *q* such that the system outputs or stores $Pr(p) > Pr(q)$ (so that its degree of confidence in *p* is not *null*), and there is also an *r* such that the system outputs or stores $Pr(r) > Pr(p)$ (so that its degree of confidence in *p* is not *maximal*).

Either way, however, we might want to say that the relevant symbolic structures also need to *play certain roles* or have certain specific *causal reverberations* within the system in order to be realizations of uncertainty in it.

This is but an instance of Fodor's (1975) requirement of causal efficacy for mental states in cognitive systems in general, in that their representational vehicles need to have some specific causal powers in order to constitute *belief* states, and different causal powers in order to constitute *intention* states, etc. (every attitude has its own causal functional profile). The same goes for uncertainty. That seems needed at least to ascribe *subjective* and not merely *epistemic* uncertainty to AI systems (the distinction was made in §1 above).

When the symbolic AI system integrates the pair <*p*, 0.9> in its model of the world, for example, we will countenance that system as having uncertainty regarding whether *p* only if that integration brings about certain consequences, but not others.

If the system is prompted to decide whether *not-p*, for example, the presence of <*p*, 0.9> in its model should cause the output of this new decision process to be <¬*p*, 0.1>, and not <¬*p*, 1>, say (¬ is the *negation* symbol here). For no agent who is uncertain about whether *p* is also sure that ¬*p*: if they are sure that *p* is *not* the case, then they are not at all uncertain about *whether p*.[10]

Related to that, we can also ascribe *merely implicit* uncertainty to symbolic AI systems—and do so *exactly because* of the potential (but not yet actualized) causal reverberations of previous uncertainties within those systems.

Say that the symbolic AI system has not only integrated <*p*, 0.9> in its model of the world, but also <*q*, 0.8>, where *p* and *q* are mutually independent according to that model. And say that the system is now prompted to decide whether $(p \wedge q)$, where $\wedge$ is symbol for *conjunction*. Supposing that this system's algorithmic information-processing abides by the principles of probability, its output will now be $<(p \wedge q), 0.72>$. We can say, then, that this system is not only *explicitly* uncertain about whether $(p \wedge q)$ *now*, but also that it was already *implicitly* uncertain about whether $(p \wedge q)$ even *before* this computation actually took place.[11]

[10] I am for the moment bracketing the possibility of *incoherence* among the system's attitudes, if only for the sake of tractability.

In symbolic AI systems, explicit uncertainty is a matter of having already tokened the relevant symbolic structures, and implicitly uncertainty is a matter of being in a position to do that without needing any new information.

### 3.2 Categorical uncertainty realized symbolically

The other type of uncertainty that can be realized in symbolic AI systems is not probabilistic in nature—it has more directly to do with *questions* that the system has, or queries that are posed to it, even if the system doesn't have any particular degree of confidence toward their answers yet.

Previously we saw that an expert system could have uncertainty regarding whether patient $a$ has disease $D$ by outputting and storing a string of symbols such as $<D(a), r>$. And now we observe that a symbolic system can also bear uncertainty on that issue by storing and processing rather an *interrogative* sentence or formula such as $?D(a)$, which translates 'Does patient $a$ have disease $D$?'. Upon getting $?D(a)$ as a query input, the system may then proceed to search in its knowledge base whether there is any set of premises from which $D(a)$ or $\neg D(a)$ can be deduced (see Sterling and Shapiro 1994). If the system can't derive either answer, and it integrates accordingly the very interrogative formula $?D(a)$ into its model of reality, then we countenance it as uncertain whether patient $a$ has disease $D$.

Here, we may assume that the target system *doesn't even support* graded judgments, or that is not equipped to deal with probabilities. It has uncertainty, but not probabilistic uncertainty. It has so to speak a *categorical* type of uncertainty. And that type of uncertainty is irreducible to credences or any other propositional attitudes—it is a fundamentally interrogative attitude (§2).

What if at time $t$ the system tokens $?D(a)$ but, given time to process the question, it *will* eventually conclude either that $D(a)$ or that $\neg D(a)$? Do we *at time t* countenance the system as uncertain about whether $a$ has $D$?

In line with the suggestion from the previous section, we could describe the situation as follows. At time $t$ the system is *explicitly* uncertain whether $a$ has $D$. It is after all tokening an interrogative sentence that expresses that very question. But that state of uncertainty is short lived, because the system will settle that question using the information and information-processing algorithms available to it *at t*. It doesn't need any new information or new bits of program to settle the question—it *already is in a position* to conclude an answer to that question at $t$. Only it hasn't done that by *t* yet. Accordingly, we can say that the system *implicitly* has an answer to its own question at *t*, though it is explicitly uncertain about it at that time.

[11] Notice that the probabilistic validity of these computations is not essential to the point. It could be the case that the system's algorithms do *not* abide by the principles of probabilities—but it still assigns a nonextreme probability to $(p \wedge q)$ in the case just described, in which case the point still holds that it is *explicitly* uncertain about whether $(p \wedge q)$ after the computation, but implicitly uncertain about whether $(p \wedge q)$ before the computation.

These are some ideas about how symbolic AI systems realize states of uncertainty. They come in two general kinds, probabilistic and categorical, both of which can be either explicit or merely implicit.

### 3.3 Levels of ascription of uncertainty to symbolic AI

But we should be careful about the *level* of interpretation from which we ascribe uncertainty to a symbolic AI system. There are at least two such levels here: the *cognitive* and the *behavioral* level.[12] In the former case, we ascribe uncertainty to the system with regard to the *program* that it is running on and the *internal computations* it performs (information-processing on symbolic representations). In the latter case, we ascribe uncertainty to the system with regard to the *observable inputs and outputs* coming in and out of it.

In the previous chapter, for example, I talked about different ways in which symbolic AI systems can be uncertain at the cognitive level of interpretation. After all, it was the fact that a symbolic structure like <*p*, 0.9> was part of a system's *internal* model of the world, and that it played a certain role in it, that made it appropriate to ascribe uncertainty about whether *p* to it. Similarly, it was the fact that a symbolic structure like *?D*(*a*) constituted the system's unanswered query and was subsequently stored and processed that made it appropriate to ascribe uncertainty about whether *D*(*a*) to it. These ascriptions are not based on the system's observable behavior, but rather on its internal operations.

But now notice that ascriptions of uncertainty may seem appropriate at the cognitive but not at the behavioral level of interpretation—and this is a form of *level split.*

An example to illustrate the phenomenon. Suppose a symbolic AI system supports the probabilistic kind of uncertainty that we saw in§3.1—say, by tokening and storing representations such as <*p*, 0.9>. And suppose additionally that, even though those are computed in accordance with the laws of probability, another part of the system's program states that, when a formula's probability is above 0.95, the system will rather output the translation of that formula into English, as opposed to a statement of its probability.

Now let *q* translate 'Quentin will make a new movie', and assume that, after computing the relevant probabilities, the system integrates <*q*, 0.96> into its internal model of reality. From this, we feel inclined to describe the system as having some uncertainty about whether Quentin will make a new movie. But we will not feel inclined to make that ascription when solely interpreting the system on the basis of its observable behavior. After all, the system will simply type or utter the sentence 'Quentin will make a new movie' to us. This system *could* have been set in a different manner, of course, so as to output something like 'The probability that Quentin will make a new movie is 0.96'—but it actually doesn't do that. It rather categorically states the English sentence translated by its internal symbol *q*.

---

[12] These are not to be confused with Marr's (1982) levels of analysis of a cognitive system. Ours are levels of *interpretation*, not *analysis* of an AI system—for example, our levels do nothing like describe the physical structure of the system (Marr's implementation level of analysis) or the problem that the system is solving (Marr's computational level of analysis).

In such cases, ascriptions of uncertainty seem to be called for at the cognitive level, but not at the behavioral level. The system *behaves as if* it had no uncertainty. But, when we look at its *internal* processing, it looks to us as if the system *does* have uncertainty.

One might dispute the accuracy of those appearances, however, and propose that if a system doesn't behave as if it were uncertain, then it just isn't uncertain, not even internally.

For shouldn't the system refrain from producing a token of 'Quentin will make a new movie' if it is uncertain whether Quentin will make a new movie? Compare to us humans: isn't it the case that we qualify or hedge our assertions when we have uncertainty, e.g., by using phrases such as 'I think…', 'I'm not a 100% sure, but…' and 'Most likely…'?[13] When I am somewhat uncertain that it is going to rain tomorrow, you don't expect me to just flatly assert that it is going to rain tomorrow (barring insincerity, coercion, etc.). You rather expect me to say that *I think* it will rain tomorrow, or that *most likely* it will, etc.

The contention may even appeal to our own principles for ascribing attitudes to symbolic AI systems from §3.1. There we said that we want the system to not only token and store the relevant symbolic structures in its internal model of the world—we also want those symbolic structure to play a certain *causal role* in order for them to constitute ways in which uncertainty is realized within the system. And now we may add that it is part of the causal role of a state of uncertainty that it causes its bearer to qualify or to hedge its assertions, or even to refrain from making them altogether. In general, a mental state's expected causal reverberations are not only internal to the system (e.g. the integration of further symbols into its internal model of the world), but also external, at the level of the system's observable behavior—including the sentences it types or utters.

We won't try to settle the issue here, but rather register the two options available at this point: *either* we treat the relevant cases as cases involving uncertainty at the purely cognitive level but not at the behavioral level of interpretation, *or* we do not treat them as cases of uncertainty to begin with (at any level). The point will be revisited in §5, where I will favor the latter option.

## 4. Uncertainty in connectionist AI systems

Level splits are also possible in connectionist AI systems. That is due to the fact that, even though they run on artificial neural networks (ANNs), they may deploy other processes or algorithms *on top* of those ANNs.

But I am going to proceed from the ground level up in this investigation. The first order of business is to establish whether and how a *single* neural network itself can realize states of uncertainty (§4.1–4.3). The issue of levels will be revisited afterwards (§4.4, §5).

---

[13] For more on the topic of hedged assertions, see Benton and van Elswyk (2020).

### 4.1 Data and model uncertainty in neural networks

There has been considerably more work on uncertainty in artificial neural networks than in symbolic AI systems. The bulk of this work has been concerned with (i) *data* uncertainty, which is uncertainty inherent in the very data used to train neural networks, and (ii) *model* uncertainty, which consists of the absence of distributed knowledge/information encoded in the weights and thresholds of the neural network itself (see Gawlikowski et al. 2023 for an overview).

As an example of data uncertainty (i), consider an ANN that is supposed to recognize whether product reviews on the internet are *sincere* or *sarcastic*. The network is trained with labelled data, and the data has been labeled by humans.[14] Suppose there was 100% agreement among human labelers regarding certain product reviews in this data. For example, reviews stating things like "I highly recommend this product" and "This product sucks" were labeled as sincere by all human labelers. But labelers were divided regarding certain, say, reviews containing phrases such as "Thanks a lot". As a result, *the very data* that is provided to the ANN in its training phase contains uncertainty as to whether those reviews are sincere or sarcastic. The data itself is ambiguous about that, for there is underdetermination of a person's intentions given the words/strings of words they typed.

As an example of model uncertainty (ii), say the ANN is supposed to categorize animals as mammals or non-mammals. Since it hasn't been trained on enough data yet, however, it lacks distributed knowledge that *bears are mammals*. The activation thresholds and weights of the connections between its units are such that, in some cases where its input layer of units registers the presence of a bear (through its activation vector), the output layer registers the presence of a mammal. But, in other cases, the output layer does *not* register the presence of a mammal, even though again the input layer registers the presence of a bear.

Here, the ANN lacks (distributed) knowledge that bears are mammals. It also lacks distributed knowledge that, if *one* bear is a mammal, then *all* of them are. This type of uncertainty consists of *lack of distributed knowledge* in the network's model of reality. An ANN would have the relevant kind of distributed knowledge just in case its activation thresholds and weights were such that, every time that the input layer registers the presence of a bear, the output layer registers the presence of a mammal. If it did that, then all bears would be mammals according to its own model.

In §1, I introduced the distinction between *epistemic* uncertainty (when the information possessed by the system is uncertain) and *subjective* uncertainty (when the *system itself* is uncertain). Now are the two types of uncertainty just introduced *subjective* or merely *epistemic* in kind? As I will show, it is not always that easy to separate these categories when it comes to neural networks.

---

[14] It is the data with which the ANN is trained that determines how the weights of the connections between its units will be set, together with the learning algorithm that it uses to update those weights—see LeCun, Bengio and Hinton (2015) for more details.

### 4.2 The relationship between epistemic and subjective uncertainty in neural networks

As characterized above, uncertainty of type (i) or *data* uncertainty is first and foremost an *epistemic* kind of uncertainty, not a subjective kind of uncertainty.

It is easy to see that. It is clearly possible for there to be uncertainty *inherent in the data* provided to a neural network even though *the network itself* is not uncertain, as per how it computes its outputs and how it behaves.[15] Data uncertainty need not bring about or lead to subjective uncertainty in an ANN, that is. Depending on how the weights of the network are set (and what the thresholds of activation for its units are), the network can 'jump to conclusions' despite the ambiguous character of its own training data. When that happens, there is data uncertainty without subjective uncertainty.

As an illustration, let us look again at one of the examples from the previous sub-section, namely, the one where the ANN was supposed to recognize whether product reviews on the internet are sincere or sarcastic.

It is perfectly possible that, even though the target ANN is trained on ambiguous labeled data (reviews labeled as sincere by some users and as sarcastic by others), it classifies some of the ambiguous reviews as sincere, thus failing to be sensitive to the fact that labelers have been divided about them. That is, regarding at least one product review *r*—for example, one containing "Thanks a lot"—the labeled data with which the ANN was trained *fails* to contain enough information to tell whether *r* is sincere or sarcastic. But *the ANN itself* takes *r* to be sincere. Its stance on the issue doesn't reflect how its total evidence or information bears on it. There is epistemic uncertainty inherent in its data regarding whether *r* is sincere or sarcastic—but the network doesn't have subjective uncertainty whether *r* is sincere or sarcastic. It has after all 'made up its mind' as to whether it is one or the other.

What about *model* uncertainty, or uncertainty of type (ii)? Is it epistemic or subjective?

The answer is that it is both at the same time. It is not possible to dissociate them here. Let us illustrate that with the second example from the previous sub-section.

Our target ANN lacks distributed knowledge whether all bears are mammals, again. But is it *thereby* uncertain whether all bears are mammals? Does the network itself have a stance of uncertainty about that, over and above the fact that its model of reality doesn't distributively encode the information that all bears are mammals?

The target network may take *some* bears to be mammals, in that its output layer generates vectors that convey the information that *x is a mammal*, given that its input layer generates vectors that convey the information that *x is a bear*, for some *x*.[16] So, regarding some particular bears, the ANN may not be uncertain whether *they* are mammals. (That *should* be enough for the ANN not to be uncertain about whether *all* bears are mammals—if not for the fact that it also lacks distributed knowledge of the fact that if a single bear is a mammal then all of them are.)

---

[15] Though note that many authors in the literature on ANNs use the qualifier 'epistemic' to refer to *model* uncertainty and not to *data* uncertainty—see for example Gawlikowski et al. (2023) and He and Jiang (2024). But by 'epistemic' they mean roughly what I mean by 'subjective' here.

[16] Whenever I write 'vectors' without qualification, I just mean *activation* vectors.

But whether the ANN itself is uncertain about whether all bears are mammals depends on what kind of output vectors it generates for those *x* such that it takes *x* to be bears but it does *not* take them to be mammals. There are two options here: either those vectors convey the information that *x is not mammal*, or they convey neither the information that *x is a mammal*, nor the information that *x is not a mammal*. But, of course, if there were a *single x* such that the ANN's input vector conveys the information that *x is a bear* and its output vector conveys the information that *x is not a mammal*, then the ANN would *not* be uncertain whether all bears are mammals. For, in that case, it would actually take it that *not all bears are mammals*. But there wouldn't be uncertainty in the ANN's *model* in that case, either. After all, the activation thresholds of this ANN and the weights of the links between its units would thereby encode the false claim that *not all bears are mammals*.

So, in order for our network to be subjectively uncertain whether all bears are mammals, it must be the case that, whenever there is an *x* such that the network takes it to be a bear but not a mammal, it simply does not take it to be a mammal, but without going so far as taking it to be a non-mammal. Correspondingly, it has to behave in exactly that way in order for there to be uncertainty regarding whether all bears are mammals *according to its model.*

That illustrates how uncertainty of type (ii), or model uncertainty, cannot be dissociated from subjective uncertainty in ANNs—in that these networks are themselves uncertain when they have model uncertainty. Any case of model uncertainty in ANNs is not only a case of epistemic uncertainty (the system's total evidence or information fails to settle the issue), but also a case of subjective uncertainty (the system itself fails to settle the issue).

Importantly, this realization of uncertainty in ANNs is also purely interrogative, in the sense that it is not reducible to propositional attitudes. The ANN is uncertain whether all bears are mammals—but this is not equivalent to its encoding any specific bit of information in a distributive manner. It is just that its model doesn't decide the issue either way (though it *may* come to decide the issue through adjustments on its weights).

Notice, furthermore, that certain architectural requirements need to be satisfied in order for subjective uncertainty to be realized in this way (through model uncertainty). For example, in order for the case we just saw to be a case of *uncertainty* whether all bears are mammals, and not a case of belief or certainty that *not all bears are mammals*, the ANN has to be able to output activation vectors that *neither* classify a bear *x* as a mammal *nor* classify it as a non-mammal, again. So its outputs cannot be single-element vectors containing only one of 1 and 0—say, <1> conveying the information that *x is mammal* and <0> conveying the information that *x is not a mammal*. For, if the ANN has that kind of architecture, then it will *always* take *x* to be a mammal or take *x* to be a non-mammal, for all *x* that it is able to categorize. But if it admits of a third activation value—<0.5>, say—then it is possible for it to neither take some *x* to be a mammal nor take it to be a non-mammal. Alternatively, it can have *two* output units instead of just one. It may be such that the output vector <1, 1> conveys the information that *x is a mammal*, for example, whereas the output vector <1, 0> conveys the information that *x is not a mammal*—and output vectors <0, 1> and <0, 0> convey *neither* of those two pieces of information.

I will not attempt to give a general characterization of the architectural requirements for a neural network to be able to be uncertain about $Q$ through model uncertainty, for any question $Q$. Such a characterization would perhaps be too abstract to be informative, so as to work for all questions that an ANN can have model uncertainty about. Further investigation into this issue is in any case left for future extensions of the present work.

In the next sub-section, I will explore yet other ways in which subjective uncertainty can be realized in neural networks.

### 4.3 Other types of subjective uncertainty realized in neural networks

Model uncertainty is a *distributive* realization of subjective uncertainty in ANNs.

That is, their own uncertainty about this or that issue is realized via the activation thresholds and weights of the connections between their units, via the number of their output units and the types of activation values that they can have, etc. In these cases, it is the *network as a whole* that is countenanced as uncertain.

To the extent that it makes sense to say that a ANN *knows* or *believes* that $p$ when it distributively encodes the information that $p$, we can also say again that the ANN is *uncertain about whether* $p$ when it neither distributively encodes the information that $p$ nor distributively encodes the information that $\neg p$, given at least that it is designed to have a stance on (among other things) whether $p$. The latter qualification is important, seeing as we don't want to countenance the network I have introduced above—which was designed to classify different animals as mammals or non-mammals—as being uncertain regarding *where you are*, for example. In order to be uncertain about where you are, a ANN needs to be designed to decide or process information about where you are in the first place (e.g. its input units must be mappable into pieces of information that are relevant to decide what is your current spatial location). So the mere fact that an ANN doesn't know where you are, or that it doesn't have any information about that, doesn't entail that it is uncertain regarding where you are.

In any case, the point now is that there are also other ways in which ANNs can realize states of uncertainty, other than in this purely distributive manner. For example, neural networks can *output* activation vectors that stand for questions—e.g. questions of the form *whether* $p$. These output activation vectors are but codes for interrogative sentences such as 'Are bears mammals?' and 'Is this review sarcastic or sincere?'. And this is also a *point-wise* realization of uncertainty, not only a distributive realization of it.

Similarly, neural networks can output probabilistic assessments and be countenanced as having different degrees of confidence in certain propositions. Neural networks, just like symbolic AI systems, are capable of having probabilistic uncertainty (see §3.1)—and even in a point-wise manner, as opposed to only distributively.

In our examples from §4.1–42 above, the ANN's outputs were *categorical*, either answering *yes* or *no* (or neither) to questions such as whether a bear is a mammal, or whether a review is sincere. But we can also arrange them so that one of the values of their vectors stands for how probable or likely something is (represented by the rest of the vector). So, for example, an output activation vector of the form <…, 0.6> may indicate that the network is 60% confident that $p$, where $p$ is the information conveyed by the first elements of that vector (here represented by the ellipsis). So we can say that the ANN is uncertain about *whether p* in this case, at least to a certain degree. Indeed, many equate or associate uncertainty in ANNs to their degrees of confidence when making predictions, categorizations, etc.[17]

The degrees of confidence of a neural network can themselves be realized in many different ways.

In the example we just saw, the proposition that the ANN has a degree of confidence toward is conveyed by part of its output activation vector. But, alternatively, the ANN may be such that *each* of its output units stands for a different proposition or proposition-type. So which degrees of confidence the ANN has on each of these propositions depends on the activation value that each of those units has in a particular case.

An example to quickly illustrate this kind of realization of uncertainty.

Let the number of output units of the target ANN be $n$. Where its input layer encodes the picture of an animal $x$, let us say that its first output unit stands for the proposition that *x is a bear*, the second one stands for the proposition that *x is a zebra*, etc. (making a total of $n$ propositions about what kind of animal $x$ is). So when the activation vector of its output layer is <0.2, 0.1, …>, for example, the ANN is 20% confident that *x is a bear*, 10% confident that *x is a zebra*, etc. The ANN will then have uncertainty about *whether x is a bear or a zebra or…* (add the other $n - 2$ kinds of animals here).

This type of probabilistic uncertainty is *point-wise* and not merely distributive, in that it is realized by a *particular* activation vector in the ANN's output layer.

### 4.4 The question of levels in connectionist AI

In the two previous sections, I explored two different ways in which subjective uncertainty can be realized in neural networks: it can be realized in a *distributive* and in a *point-wise* manner.

But here it may be pointed out that these types *really are* realizations of subjective uncertainty only if they have certain causal powers, again. This is the same requirement as the one we saw for symbolic AI systems in §3.1, though applied to a different architecture—the requirement that a state of subjective uncertainty must have a certain functional profile, or play a certain role within the system on which it is realized.

---

[17] This way of talking about neural networks (as if they had different degrees of confidence on different claims) is very natural and it appears even in introductions to AI/Machine Learning—see for example Ch. 5–6 of Mitchell (2019).

Do any of the realizations of uncertainty we saw above have causal powers characteristic of uncertainty?

Consider *distributive* realizations of uncertainty first. It is more or less clear that, by virtue of being distributively uncertain, the network itself has causal dispositions and powers characteristic of uncertainty. In our example of a distributive realization of uncertainty, the ANN was uncertain *whether all bears are mammals*—which is realized through the weights of the links between the units of the network (and their activation thresholds). *As a result* of being that way, however, there are particular cases where the ANN takes an object to be a bear, but it is uncertain whether *it* is a mammal, in that it doesn't take it to be a mammal and it doesn't take it to be a non-mammal, either. That is a causal relation: the ANN's being distributively uncertain about *whether all bear are mammals* causes it to be uncertain about whether some particular bears are mammals. And this is indeed a causal power characteristic of uncertainty—uncertainty regarding a general question will dispose the system to be uncertain about more particular questions, too.[18]

So *distributive* realizations of uncertainty do have causal powers characteristic of uncertainty, and these are powers to bring about further consequences *within the network itself* when it takes different inputs.

But the same cannot be said of point-wise realizations of uncertainty in neural networks. Here, the uncertainty ends in the network itself, without bringing anything else about in it.

Point-wise uncertainty is after all realized through *output* activation vectors of the network. But what happens *to* the network as a consequence of those outputs? The network itself doesn't do anything *with them*.

And here we are immediately led to think of levels of interpretation of connectionist systems. For, in order for a point-wise state to have causal powers and a functional profile characteristic of uncertainty, the network itself has to be *part of a larger system*—so that this system as a whole can do something with the network's outputs. And not just do any old thing, but specific things with those outputs. For that which the cognitive system as a whole does with the ANNs outputs can be not specific enough or even the exact opposite of what is expected from a state of uncertainty.

Here are two examples of ANNs embedded in a larger system to illustrate phenomenon (the second one of which is analogous to what we saw could also happen in symbolic AI systems earlier in §3.3).

First, consider the typical system of an artificial neural network *plus* a learning algorithm that is supposed to *correct* or *fine-tune* that network. Say the algorithm is a *back-propagation* algorithm (LeCun, Bengio and Hinton 2015). The algorithm will calculate the difference between the ANN's *actual* output vector and the *desired* output vector and use that difference (if any) to modify the weights of the connections between units in the network (to propagate the error back to the network). Clearly, even if the output vector looks like a realization of uncertainty in the network, *that which the larger system does with it* does not apply *specifically* to states of uncertainty—it applies to all possible output vectors indiscriminately.

---

[18] In the present case, we can think of it roughly as follows. If the ANN is disposed to take *every* bear to be a mammal, then it is not uncertain *whether all bears are mammals*. Contrapositively, if it is uncertain *whether all bears are mammals*, then it is *not* disposed to take every bear to be a mammal. It is rather disposed not to take some bears to be mammals.

The second example is as follows. Say again the ANN's output activation vector is <…, 0.85>, and that means that it is 85% confident that $p$ (where $p$ is again the information conveyed by the rest of the activation vector, represented here by the ellipsis). But say that this ANN is coupled to a larger cognitive system, and that its output is processed by an overarching algorithm within that system which decides that, if the target ANN is more than 80% confident that $p$, then $p$. That is, the whole cognitive system treats $p$ as true, given that the ANN's subjective probability for $p$ is bigger than 0.8.

In this case, is the ANN's output activation vector really a realization of uncertainty?

After all, the whole system doesn't *behave* as if it were uncertain that $p$ when the ANN's output activation vector is <…, 0.85>. Quite the opposite, the whole system acts as if it were certain that $p$! And this may be the case not only at the level of *behavior*, but also *cognitively*, at the level of internal information processing. The system may draw further inferences from $p$, for example, or build possible models of the world/environment that rely on the presupposition that $p$.

The worry here is that the ANN's output vector *doesn't play the role* of a state of uncertainty. It doesn't have causal powers characteristic of uncertainty within the larger cognitive system in which it is processed—therefore it is *not* a realization of uncertainty.

But saying that the target output vector is not at a realization of uncertainty at all is once again not the *only* option. As before, the alternative is to say that there is a kind of level-split in ascriptions of uncertainty here, too: *the network itself* is uncertain about whether $p$, but the whole cognitive system is not uncertain whether $p$.

In the next section, I assess which option is best, both when symbolic and connectionist AI systems are concerned.

## 5. The general problem of level splits

In both, the case of symbolic AI (§3.3) and that of connectionist AI (§4.4), we are faced with a puzzle of sorts. In each case, the puzzle stems from the observation that a larger system may be composed out of a sub-system that seems uncertain about something, in that it only has nonextreme degrees of confidence on the topic, even though the larger system seems not to be uncertain about it, in that it behaves/cognizes as if it had certainty. So, is uncertainty realized in the larger system or not?

One option we saw is to deny that the sub-system's states are realizations of uncertainty at all, in that they don't *play the role* that states of uncertainty are supposed to play within the larger system, *qua* states of uncertainty. Call that the First Solution.

The other option is to grant that the sub-system's states are realizations of uncertainty, and that they are states of uncertainty realized within the larger system, too. To that extent, the larger system is uncertain. But then the larger systems somehow manages to *ignore* its own uncertainty. Call that the Second Solution.

The Second Solution may seem better at first.

A cost of *not* countenancing the target sub-system states as realizations of uncertainty, after all, is that we thereby commit ourselves to the intuitively incorrect claim that the larger system is *not* failing to be responsive to its own uncertainty in the cases we saw. For there is no such uncertainty for it to fail to be responsive to. For example, when the subsystem outputs a nonextreme degree of confidence that *p*, but the whole system behaves and cognizes as if it were certain that *p*, the system is *not* failing to be responsive to its own uncertainty about whether *p*. Now, if we were to say that a state of uncertainty *is* realized in the system in this case, we would thereby be able to say that the system is defective, in that it is not respecting its own uncertainty. But if we don't ascribe uncertainty to the larger system, then we cannot criticize it or find problems with it on those grounds—as it is more or less clear we are entitled to.

It is in view of such considerations that it may seem preferable to say that, in the cases we looked at above, the larger cognitive system *does* realize states of uncertainty—though it again fails to be properly responsive to them, or it ignores its own uncertainty (Second Solution).

But this purported solution is not free of problems, either. There are at least two main issues with it.

The first one is that there may also be cases where a larger system is composed out of *more than one subsystem*—e.g. two neural networks, or two separate symbolic processing algorithms—such that they both deal with the same issue, though the system's overarching algorithm gives priority to one of those subsystems over the other when they fail to mutually cohere. Now suppose we are interpreting such a composite system, and that one of its sub-systems realizes a state of uncertainty (according at least to the Second Solution). Does it follow that the larger system is also uncertain? Not necessarily, for it may happen that its *other* subsystem has certainty in one of the answers to the question that the other subsystem is uncertain about and, in such cases, the overarching algorithm gives priority to the system that is certain. In this case, it would seem that a state of uncertainty is *not at all* realized by the larger system. But isn't the defender of the second solution committed to saying about this case that, similarly, because the subsystem is uncertain, the larger system is uncertain—though it fails to be responsive to its own uncertainty?

That clearly calls for clarification of the criteria for ascriptions of uncertainty to a system on the part of defenders of the Second Solution. They have to explain why *in some cases* the larger system inherits the uncertainty of its sub-system, though fails to be responsive to it, whereas in *other* cases the larger system does not inherit the uncertainty of its sub-systems.

The second issue is the following. It seems that we should not countenance neural networks as uncertain when they jump to conclusions—that is, when the ANNs themselves conclude or categorically output answers that are merely probable. But it is not clear that the Second Solution coheres with this verdict.

ANNs are well known for their occasional *overconfidence* (see Guo et al. 2017, Ao et al. 2023). They may predict that it will rain by outputting an activation vector such as <1>, for example, merely because the *threshold* of activation of that output unit is somewhat low. Let us say, for example, that its total input registers the probability of rain. If its total input is bigger than or equal to 0.6, its output activation value is <1> and, if its total input is smaller than 0.6, its output activation value is <0>.

Regarding such a network, a defender of the second solution will point out that their solution does not entail that the ANN *is* uncertain about whether it will rain, though it shouldn't be. They will point out that their view is compatible with saying that the ANN *is not* uncertain at all about whether it will rain, in that it is certain that it will rain. Accordingly, they can add that this ANN is not respecting the *epistemic* uncertainty of rain. There is epistemic uncertainty about that, but no subjective uncertainty. This is at least a theoretical option that is open to those who favor the second solution and reject the first one. There is after all no inconsistency between saying that the ANN we just saw is not uncertain about whether it will rain, on the one hand, and saying that a larger system built out of a ANN inherits that ANN's very uncertainties.

That sounds correct but, under the Second Solution, it is unclear again what are the criteria for when to ascribe uncertainty to these systems when there is uncertainty in their subsystems. For why shouldn't we say, rather, that the ANN we just saw doesn't respect its own uncertainty, too, just like the larger system didn't respect its own uncertainty in the case of a level-split, according to the Second Solution?

Contrast all that to the First Solution and how straightforward it is: it says that the system as a whole has no subjective uncertainty in *either* of these cases. The states realized by its sub-systems are *not* realizations of uncertainty, because they do not play the role of states of uncertainty within the larger system in which they are embedded. For example, those states do not cause the larger system to hesitate when making decisions that hinge on whether *p*. The system rather acts and cognizes as if it were certain that *p*.

Clearly, the First Solution is simpler than the Second Solution—and that is a reason to prefer it, at least when it comes addressing the issue of levels of ascription of uncertainty to these AI systems.

## 6. Concluding remarks

Let me sum up the results of this investigation so far.

There are at least two ways in which symbolic AI systems can realize states of uncertainty. First, they can realize uncertainty by realizing *nonextreme degrees of confidence*, which are in turn realized through symbolic structures that assign numeric or comparative probabilities to different propositions. I called that probabilistic uncertainty. Second, they can realize uncertainty through *interrogative* formulas or sentences—symbolic structures that pose a question, as opposed to convey an answer to it. I called that categorical uncertainty. This type of uncertainty is fundamentally interrogative, irreducible to degrees of confidence or any other propositional attitudes.

In connectionist systems, too, there are probabilistic and categorical realizations of uncertainty. In the case of ANNs, however, these states can be realized in a purely *distributive* manner. For example, a network can be uncertain whether a certain generalization holds by virtue of how they weights of its connections and the thresholds of its units are set, in such a way that it is disposed to endorse *some* but *not all* instances of those generalizations. That said, ANNs may also realize uncertainty in a *point-wise* manner, namely, through the activation vectors of its output layer. Those vectors may either be realizations of nonextreme degrees of confidence (output vectors are probability estimates), or realizations of irreducible interrogative attitudes (output vectors express questions).

But I argued that the latter are point-wise realizations *of subjective uncertainty* just in case the target ANNs are embedded in larger cognitive systems, and their outputs play a role or have causal reverberations within that system that are characteristic of states of uncertainty. This coheres with the fact that, when we demand that AI systems should be more uncertain, or when we complain that they aren't, what we want them to do is really to behave like an uncertain agent—not to just have internal states of a certain kind.

The latter point is also connected to a more general issue with ascriptions of uncertainty to AI systems of all kinds. In cases of level-splits, when a subsystem seems to be uncertain but the larger system that embeds it does not, I have argued, it is preferable to countenance the whole system as *not* realizing a state of uncertainty at all (First Solution). The other option (Second Solution) is riddled with avoidable problems.

This inquiry was divided in two parts, one concerned with symbolic AI systems and the other one concerned with connectionist systems. The conclusions drawn about each of these types, together with the conclusions drawn about level-splits, give us also answers to questions about hybrid systems—they can realize uncertainty in *all of the above* ways. But a more thorough analysis of uncertainty in hybrid systems, and also a comprehensive assessment of different case studies in the vast field of AI, is left for future extensions of the present work.

## References

Anderson, John Robert (1976). *Language, memory, and thought*, Mahwah, NJ: Lawrence Erlbaum Associates.

Ao, Shuang, Stefan Rueger and Advaith Siddharthan (2023). 'Two Sides of Miscalibration: Identifying Over and Under-Confidence Prediction for Network Calibration,' arxiv preprint arXiv:2308.03172.

Benton, Matthew and Peter van Elswyk (2020). 'Hedged Assertion,' in S. Goldberg (ed.) *The Oxford Handbook of Assertion*, Oxford: Oxford University Press, pp. 245–263.

Bermúdez, José Luis (2023). *Cognitive Science: An Introduction to the Science of the Mind* (4th ed.), Cambridge: Cambridge University Press.

Chalmers, David (2025). 'Propositional Interpretability in Artificial Intelligence,' arxiv preprint arXiv:2501.15740.
Dennett, Daniel (1971). 'Intentional Systems,' *Journal of Philosophy* 68(4): 87–106.
Friedman, Jane (2017). 'Why suspend judging?', *Noûs* 51(2): 302–326.
Guo, Chuan, Geoff Pleiss, Yu Sun and Killian Weinberger (2017). 'On Calibration of Modern Neural Networks,' arxiv preprint arXiv:1706.04599.
He, Wenchong and Zhe Jiang (2023). 'A Survey on Uncertainty Quantification for Deep Learning: An Uncertainty Source's Perspective,' arxiv preprint arXiv:2302.13425v3.
Jackson, Elizabeth (2020). 'The Relationship Between Belief and Credence,' *Philosophy Compass* 15(6): 1–13.
LeCun, Yann, Yoshua Bengio and Geoffrey Hinton. 'Deep learning,' *Nature* 521(7553): 436–444.
Lewis, David. 'Psychophysical and Theoretical Identifications,' *Australasian Journal of Philosophy* 50(3): 249–258.
Marr, David (1982). *Vision: A Computational Investigation into the Human Representation and Processing of Visual Information*, San Francisco: W. H. Freeman.
Mitchell, Melanie (2019). *Artificial Intelligence: A Guide for Thinking Humans*, New York: Farrar, Straus, and Giroux.
Newell, Allen and Herbert Simon (1972). *Human problem solving*, Englewood Cliffs NJ: Prentice-Hall.
Pearl, Judea (1988). 'Probabilistic Reasoning in Intelligent Systems: Networks of Plausible Inference,' San Francisco CA: Morgan Kaufmann Publishers Inc.
Putnam, Hilary (1975). 'The Nature of Mental States,' in *Mind, Language, and Reality: Philosophical Papers*, (Volume 2), Cambridge: Cambridge University Press, pp. 429–440.
Rosa, Luis (2024). *Rules for the Inquiring Mind*, New York: Routledge.
Shortliffe, Edward and Bruce Buchanan (1975). 'A model of inexact reasoning in medicine,' *Mathematical Biosciences* 23(3–4): 351–379.
Stanley, Jason (2008). "Knowledge and certainty," Philosophical Issues 18 (1): 35-57.
Sterling, Leon and Ehud Shapiro (1994). *The Art of Prolog*, Cambridge MA: MIT Press.
Sullivan, Rob and Nelly Elsayed (2024). 'Can Large Language Models Act as Symbolic Reasoners?' arxiv preprint arXiv:2410.21490.
Willard-Kyle, Christopher (2023). 'The Knowledge Norm for Inquiry', *Journal of Philosophy* 120(11): 615–640.
Xiong, Haoyi, Zhiyuan Wang, Xuhong Li, Jiang Bian, Zeke Xie, Shahid Mumtaz, and Laura E. Barnes (2024). 'Converging Paradigms: The Synergy of Symbolic and Connectionist AI in LLM-Empowered Autonomous Agents,' arxiv preprint arXiv:2407.08516.